# PromptGuard at BLP-2025 Task 1: A Few-Shot Classification Framework Using Majority Voting and Keyword Similarity for Bengali Hate Speech Detection

**Rakib Hossan**
Bangladesh University of Business and Technology, Bangladesh
rakib911hossan@gmail.com

**Shubhashis Roy Dipta**
University of Maryland, Baltimore County, USA
sroydip1@umbc.edu

## Abstract

***Content Warning:*** *This paper contains content that some readers may find inappropriate or disturbing.*

The BLP-2025 Task 1A requires Bengali hate speech classification into six categories. Traditional supervised approaches need extensive labeled datasets that are expensive for low-resource languages. We developed PromptGuard, a few-shot framework combining chi-square statistical analysis for keyword extraction with adaptive majority voting for decision-making. We explore statistical keyword selection versus random approaches and adaptive voting mechanisms that extend classification based on consensus quality. Chi-square keywords provide consistent improvements across categories, while adaptive voting benefits ambiguous cases requiring extended classification rounds. PromptGuard achieves 67.61 micro-F1, outperforming n-gram baselines (60.75) and random approaches (14.65). Ablation studies confirm chi-square based keywords show the most consistent impact across all categories. [1]

## 1 Introduction

The widespread presence of hate speech on online platforms threatens user safety worldwide. While Natural Language Processing has advanced through transformer models (Vaswani et al., 2017), pre-trained models like BERT (Devlin et al., 2019), and large language models (Brown et al., 2020), Bengali faces unique challenges in automated content moderation due to its rich morphology and culturally specific expressions of toxicity. Additionally, the field suffers from a lack of large-scale annotated datasets (Romim et al., 2021; Das and Mukherjee, 2023). The BLP-2025 Task 1A (Hasan et al., 2025a,b) requires classifying Bengali text into six categories: none, sexism, abusive, profane, religious hate, and political hate. This task comes with some significant challenges: severe class imbalance with limited training data for Bengali hate speech detection (Faria et al., 2024), the scarcity of comprehensive datasets (Al Maruf et al., 2024), and resource constraints common to low-resource language processing (Magueresse et al., 2020).

Although Bengali hate speech detection has been explored in prior work, most systems fail to generalize under data imbalance and overlapping linguistic cues between categories such as abusive and profane (Romim et al., 2021; Das and Mukherjee, 2023). We introduce PromptGuard, a statistically grounded few-shot framework combining: (1) chi-square analysis for discriminative keyword extraction (Azzahra et al., 2021), (2) systematic two-phase example selection inspired by few-shot learning principles (Brown et al., 2020), and (3) adaptive majority voting (Dietterich, 2000) that extends decisions when consensus is unclear. Our approach demonstrates that few-shot methods combining statistical feature selection and adaptive voting can achieve competitive performance without requiring extensive labeled datasets.

To summarize, our contributions are: (1) novel integration of chi-square feature selection with few-shot prompting, (2) adaptive majority voting with dynamic decision extension, and (3) comprehensive ablation studies validating each component's effectiveness.

## 2 PromptGuard

PromptGuard combines chi-square keyword extraction, adaptive example selection, and majority voting with consensus extension to achieve robust Bengali hate speech classification without extensive labeled datasets. PromptGuard combines chi-square keyword extraction, adaptive example selection, and majority voting with consensus ex-tension to achieve robust Bengali hate speech classification without extensive labeled datasets. The framework operates through an iterative decision-making process (illustrated in Fig. 1): it begins by running 3 parallel classification attempts, each using different example sets. If these initial attempts produce a clear majority vote (>50% agreement), the winning classification is returned immediately. However, when no clear consensus emerges, the system adaptively extends the voting process by adding additional classification rounds with randomly selected examples. This extension continues iteratively until either a clear majority is achieved or a maximum of 10 total turns is reached, at which point the system selects from among the tied labels. This adaptive mechanism proves particularly valuable for ambiguous cases where the initial classification attempts disagree. The complete method pipeline is illustrated in Fig. 1.

[1] https://github.com/Rakib911Hossan/PromptGuard

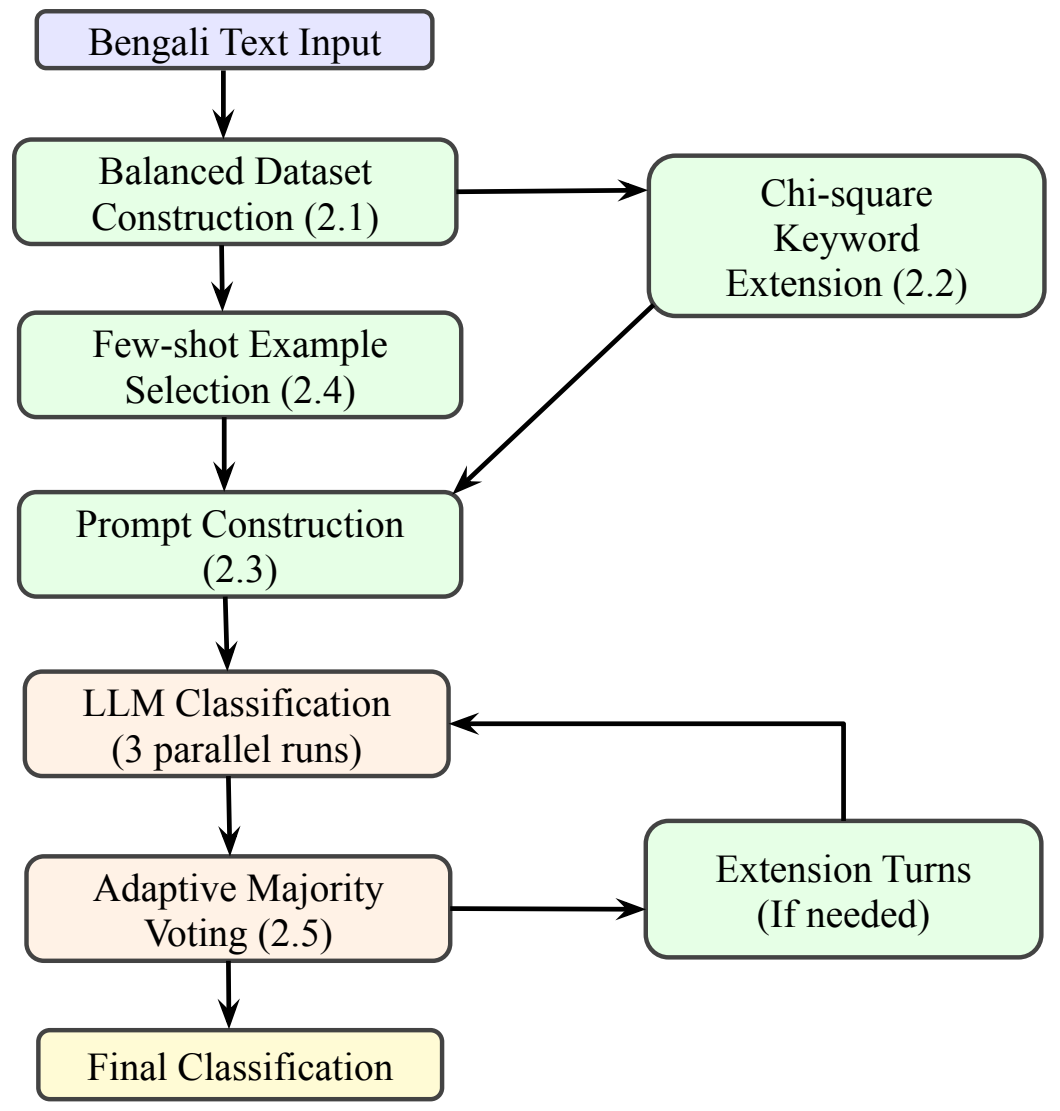

Figure 1: Overall architecture of PromptGuard framework. The system processes Bengali text through balanced dataset construction, extracts discriminative keywords using chi-square analysis, constructs prompts with selected examples, performs parallel LLM classifications, and applies adaptive majority voting to produce the final hate speech category classification.

### 2.1 Balanced Dataset Construction

The first step creates a balanced training pool by sampling 120 examples per category, resulting in 720 examples across 6 categories. This size was constrained by the smallest category (sexism) to ensure balanced representation without data augmentation. Random sampling with a fixed seed ensures reproducibility and prevents category bias.

| Category | Top-2 Keywords | $\chi^2$ Score |
|---|---|---|
| Profane | বাল | 2980.27 |
| | মাগির | 1559.10 |
| Political Hate | ভোট | 1738.07 |
| | বিএনপি | 1534.63 |
| Religious Hate | মুসলিম | 1378.39 |
| | হিন্দু | 1280.53 |
| Sexism | নারী | 871.51 |
| | পরকিয়া | 801.57 |

Table 1: Top-2 discriminative keywords per hate speech category ranked by chi-square scores. Profane category shows the highest statistical association ($\chi^2$ = 2980.27), while sexism exhibits the lowest scores among hate categories ($\chi^2$ = 871.51).

### 2.2 Statistical Keyword Extraction using Chi-Square Testing

We have used the chi-square ($\chi^2$) to identify words most statistically associated with each hate speech category (Forman, 2003), ensuring genuine discriminative power over arbitrary selection. Table 1 presents the highest-scoring keywords for each category based on chi-square analysis. The preprocessing pipeline applies Unicode filtering, minimum document frequency of 5, and maximum of 95% to retain discriminative Bengali vocabulary.

**Manual Keyword Refinement** The statistically extracted keywords undergo manual filtering for cultural sensitivity while maintaining balanced representation across categories. The refined keyword sets include:

- **Abusive:** দালাল, টিভি, ফালতু, চোর, মিথ্যা, পাগল, জুতা, লজ্জা, আমিন
- **Profane:** বাল, মাগি, খানকি, বেশ্যা, দফা, বাচ্চা, সালা, শালা, মাদারচোদ, কুত্তা, জারজ, পোলা, শুয়োর
- **Religious Hate:** মুসলিম, হিন্দু, ইহুদি, মুসলমান, গজব, ধর্ম, ইসলাম, কাফের, মসজিদ, ধর্মীয়, মোল্লা, আল্লাহ
- **Political Hate:** ভোট, বিএনপি, আওয়ামী, লীগ, সরকার, নির্বাচন, হাসিনা, অবৈধ, জনগণ, পার্টি, দল, চোর, রাজনীতি

| Model | Random | Majority | n-gram | **$\mathcal{P}$rompt$\mathcal{G}$uard** | Rank-1 | Rank-2 | Rank-3 |
|---|---|---|---|---|---|---|---|
| micro-f1 | 14.65 | 57.60 | 60.75 | 67.61 | 73.62 | 73.45 | 73.40 |

Table 2: $\mathcal{P}$rompt$\mathcal{G}$uard performance compared to baselines and top-performing systems on 10,000 test instances. While outperforming simple baselines by 7-53 micro-F1 points, a 6-point gap remains with leading approaches.

- **Sexism:** নারী, পরকিয়া, মহিলা, পুরুষ, হিজরা, বিয়ে, লিঙ্গ, হোটেল, মেয়ে, বেডা, আবাসিক

### 2.3 Enhanced Prompt Engineering

The method uses a structured prompt (§A.1) template that combines statistical keywords with few-shot examples to improve classification accuracy. Specifically, the prompt integrates three key components: (1) Examples for each category, (2) Keywords from each category, (3) step-by-step reasoning guidelines.

### 2.4 Dynamic Few-Shot Learning Strategy

The example selection method uses two phases to balance diversity, as sample selection strategies significantly impact few-shot learning performance (Pecher et al., 2024). In the first phase (3 turns), examples are selected sequentially from a pool of 120 instances without reuse, using 20 examples per category per turn. For future turns, if needed, the system extends voting for up to 10 additional turns using random selection of examples to provide fresh perspectives on ambiguous cases. If consensus remains unclear after maximum iterations, the system selects one random winner from the tied ones (the label with the highest vote count) as the final decision.

### 2.5 Robust Majority Voting with Adaptive Extension

The voting mechanism operates in two phases building on self-consistency approaches (Wang et al., 2022):

**Initial Phase:** Execute $n_0 = 3$ parallel voting turns $\{V_1, V_2, V_3\}$, each using distinct example sets $E_i$ where $E_i \cap E_j = \emptyset$ for $i \neq j$.

**Adaptive Extension:** If no clear majority emerges, iteratively add voting turns $V_{3+k}$ for $k = 1, 2, \ldots, 10$, where each $V_{3+k}$ uses freshly sampled examples $E_{3+k} \sim \mathcal{D}$. The process terminates when $\max_c |S_c| > \frac{1}{2}|S|$ or maximum iterations are reached, where $S_c$ represents votes for candidate $c$.

## 3 Results

### 3.1 Experimental Setup

We use `Qwen/Qwen3-32B` with temperature=0 and parallel processing for efficiency. Following the official shared task, we have used `micro-f1` as the main evaluation. To give more insights into the result, we have also provided confusion matrix with fine-grained analysis for each metric. All experiments were conducted on a single NVIDIA L40S (46 GB) GPU, and the full evaluation on the test set required approximately four hours.

### 3.2 Discussion

As a baseline, we have provided the random, majority and n-gram baseline. To compare with other participants' work, we have also reported the $1^{st}$, $2^{nd}$ and $3^{rd}$ results from the official leaderboard. We have reported the results on the Table 2. While $\mathcal{P}$rompt$\mathcal{G}$uard achieves better scores than the original baselines, still it lags behind compared to the Rank-{1,2,3} models.

To perform a fine-grained analysis of class-wise performance, we present the confusion matrix in Fig. 2. The breakdown reveals that the model exhibits a bias toward labeling instances as non-hate, more than any other category. The poorest performance is observed in the "Abusive" category. One possible reason for this is that many of the keywords associated with the Abusive class, i.e., টিভি, মিথ্যা, পাগল, জুতা, লজ্জা, আমিন – can also appear in benign, non-hateful contexts. This semantic overlap makes the classification of the "Abusive" category particularly challenging.

## 4 Ablation Studies

Due to the high computation and time needed to run on the whole test set (10k), we sampled a balanced subset from the test set. In the test set, the sexism category has the lowest number of examples (29). We therefore sampled a balanced subset of 174 instances (29 per label).

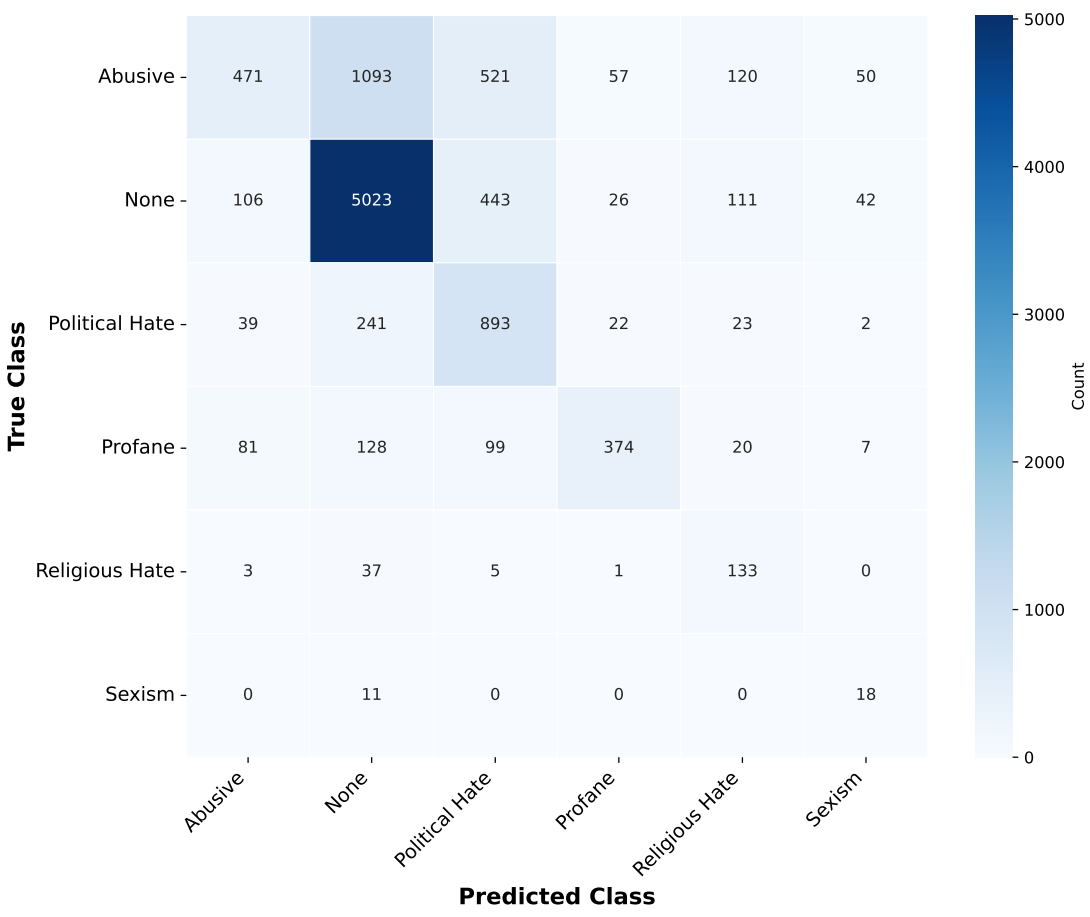

Figure 2: Per-category breakdown reveals uneven performance across hate speech types. Non-hate content is classified most accurately, while abusive language detection proves most challenging.

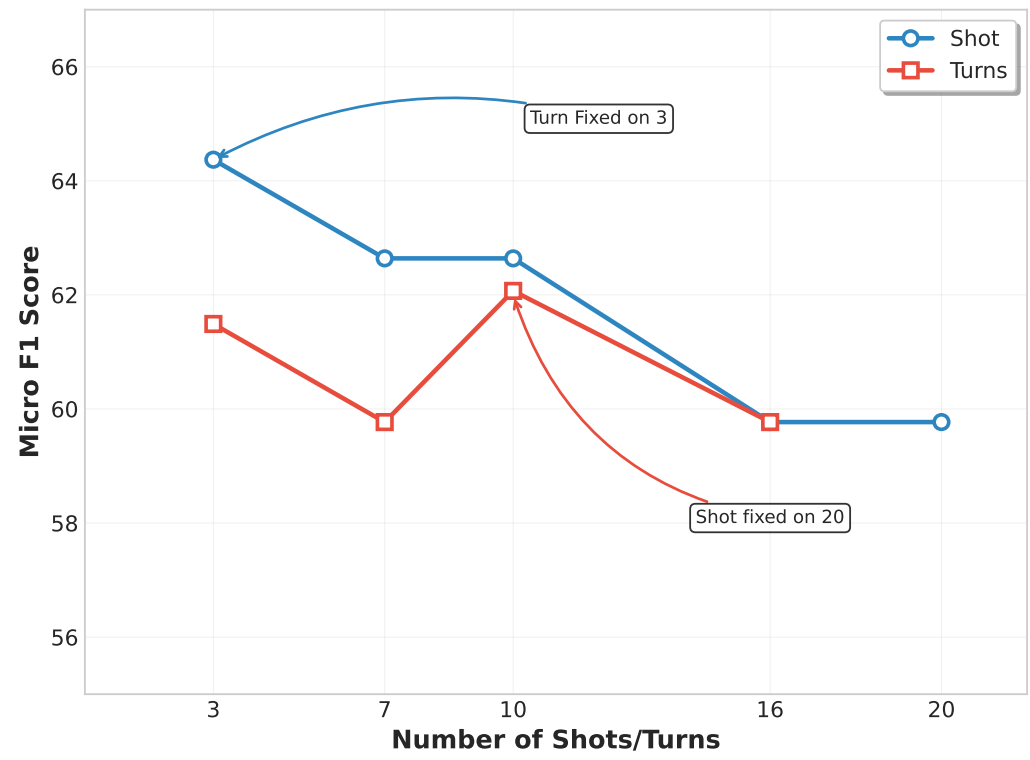

Figure 3: Impact of varying the number of examples and turns on final evaluation performance. We observe a negative correlation with the number of examples, likely attributable to the "lost-in-the-middle" problem (Liu et al., 2023).

### 4.1 Impact of Similar Keywords

As described in §A, our final model utilizes a well-crafted prompt that includes both in-context examples and targeted keywords. To assess the contribution of these keywords, we compare performance against a baseline using a basic prompt without the keywords. Both prompt versions are provided in §A.

On our sampled test dataset, the keyword-free prompt achieves a micro-F1 score of 57.47, whereas the prompt with keywords reaches 59.77, highlighting the effectiveness and importance of including targeted keywords in the prompt design.

### 4.2 Impact of Shots & Turns

In our original configuration, we used 20 examples per label in the prompt and 3 initial turns. To evaluate the impact of these two parameters, we conducted a controlled analysis by varying one while keeping the other fixed. Specifically, we experimented with {3, 7, 10, 16, 20} shots and {3, 7, 10, 16} turns. The results of this ablation are presented in Fig. 3.

The results indicate that the best performance is achieved with 3 shots and 3 turns. We hypothesize that this is due to the "lost-in-the-middle" effect in LLMs (Liu et al., 2023), where too many in-context examples can reduce the model's focus on relevant inputs. In contrast, varying the number of turns has minimal impact on performance. This aligns with our expectations, as the initial turns are primarily used to identify a clear winner; additional turns are also invoked if the earlier ones fail to produce a confident outcome.

### 4.3 Impact of Different Models

In our primary results, we used the `Qwen3-32B` model. To evaluate the impact of model architecture and size, we further ran our method on two families of models: Qwen3 (Yang et al., 2025) and GPT-OSS (OpenAI et al., 2025), across varying model sizes. The comparative results are presented in Fig. 4.

The findings show a clear positive correlation between model size and micro-F1 score within the Qwen3 family, indicating that larger models yield better performance. In contrast, the GPT-OSS models display relatively smaller performance variation across sizes and consistently lag behind the best-performing Qwen models.

## 5 Related Works

**Bangla Hate Speech Detection.** Karim et al. (2021) achieved F1-scores of 78-91% using transformer ensembles, while Jahan et al. (2022) developed domain-specific BanglaHateBERT with 1.5M Bengali posts. Raihan et al. (2023) addressed code-mixed content through cross-lingual transformers for transliterated text.

**LLM in Bengali.** Hasan et al. (2023); Roy Dipta and Vallurupalli (2024) showed that monolingual transformers outperform general-purpose LLMs (e.g., Flan-T5, GPT-4) on zero/few-shot Bangla sentiment tasks, while Wang et al. (2025) improved LLM performance through multilingual prompting enriched with cultural cues. To address reasoning limitations in LLMs, Colelough and Regli (2025) identified major gaps in explainability and meta-cognition across Neuro-Symbolic AI literature. In parallel, Kowsher et al. (2022) introduced Bangla-BERT with language-specific pretraining, outperforming prior models by up to 5.3%. Gao et al. (2025) tackled prompt design challenges with MAPS, an automated framework achieving higher coverage via diversity-guided search.

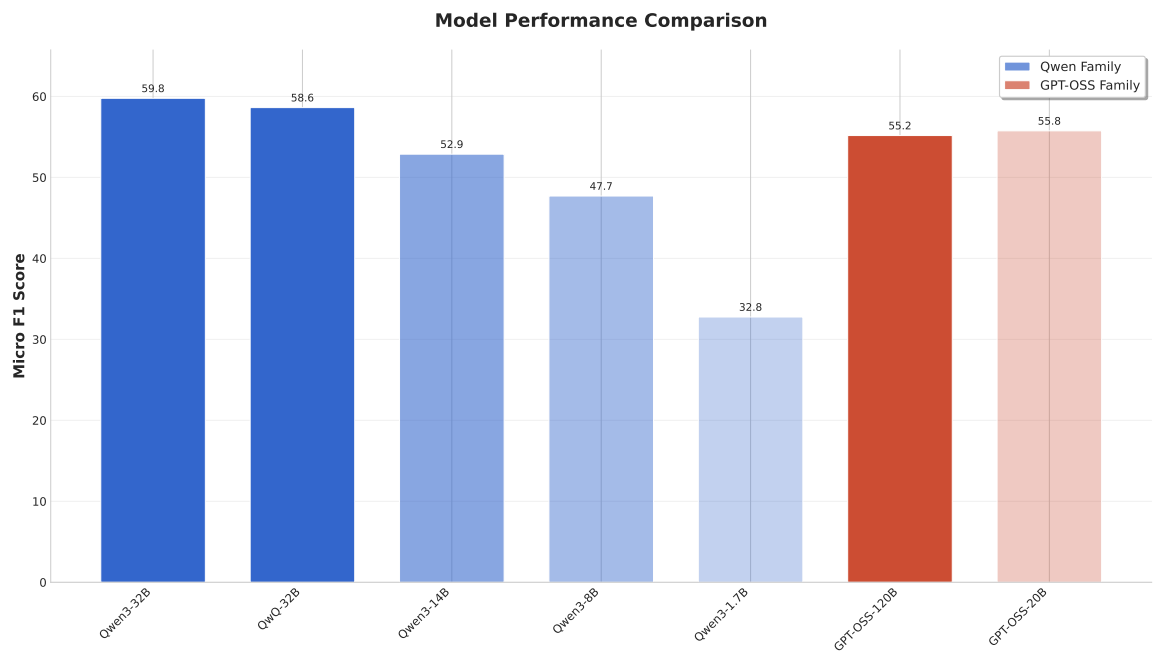

Figure 4: Performance comparison across model architectures and sizes. Qwen3 models show clear scaling benefits with size, while GPT-OSS models exhibit consistent but lower performance across different scales.

**Social Media Challenges.** Guo et al. (2024) categorized LLM biases and proposed mitigation strategies, while Natsir et al. (2023) examined dynamic language evolution in social media, identifying shifts in multilingual adaptation.

## 6 Conclusion

We present $\mathcal{P}$rompt$\mathcal{G}$uard, a few-shot classification framework that addresses Bengali hate speech detection through statistical feature selection and adaptive decision-making. Our approach combines chi-square-based keyword extraction with majority voting to achieve robust classification. Our work demonstrates the value of integrating statistical foundations with few-shot learning for low-resource language tasks. Future directions include exploring advanced feature selection methods to improve few-shot hate speech detection.

# A Prompts

## A.1 Prompt with Keywords

Following is the prompt that we have used with our final method.

You are an AI language model specialized in detecting hate speech in Bengali. Your task is to classify a given Bengali sentence into one of six categories: none, sexism, abusive, profane, religious hate, or political hate.

First, review these examples of sentences for each category:
<examples>
{{EXAMPLES}}
<examples>

Now, consider these common words associated with each category. Note that the presence of these words doesn't guarantee classification into that category, but they can be helpful indicators:

<category_keywords>
abusive: দালাল, টিভি, ফালতু, চোর, মিথ্যা, পাগল, জুতা, লজ্জা, আমিন
profane: বাল, মাগি, খানকি, বেশ্যা, দফা, বাচ্চা, সালা, শালা, মাদারচোদ, কুত্তা, জারজ, পোলা, শুয়োর
religious hate: মুসলিম, হিন্দু, ইহুদি, মুসলমান, গজব, ধর্ম, ইসলাম, কাফের, মসজিদ, ধর্মীয়, মোল্লা, আল্লাহ
political hate: ভোট, বিএনপি, আওয়ামী, লীগ, সরকার, নির্বাচন, হাসিনা, অবৈধ, জনগণ, পার্টি, দল, চোর, রাজনীতি
sexism: নারী, পরকিয়া, মহিলা, পুরুষ, হিজরা, বিয়ে, লিঙ্গ, হোটেল, মেয়ে, বেডা, আবাসিক
<category_keywords>

Here's the Bengali sentence you need to classify:
<input_sentence>
{{INPUT_SENTENCE}}
<input_sentence>

Before making your final classification, analyze the sentence in detail. Consider the following:
1. Compare the sentence to the provided examples.
2. Examine the tone, specific words used, and overall context.
3. Check if any words from the category_keywords are present and relevant.
4. For each category (none, sexism, abusive, profane, religious hate, political hate):
    - List evidence for classifying the sentence into this category.
    - List evidence against classifying the sentence into this category.
5. Summarize your findings and explain your final decision.

Your final output should be the category classification. Use only one of these exact category names: none, sexism, abusive, profane, religious hate, or political hate.

Provide your classification inside <classification> tags.

## A.2 Basic Prompt

Following is the prompt that we used during the ablation study of similar keywords.

You are an AI language model specialized in detecting hate speech in Bengali. Your task is to classify a given Bengali sentence into one of six categories: none, sexism, abusive, profane, religious hate, or political hate.

First, review these examples of sentences for each category:
<examples>
{{EXAMPLES}}
<examples>

Here's the Bengali sentence you need to classify:
<input_sentence>
{{INPUT_SENTENCE}}
</input_sentence>

Before making your final classification, analyze the sentence in detail. Consider the following:
1. Compare the sentence to the provided examples.
2. Examine the tone, specific words used, and overall context.
3. For each category (none, sexism, abusive, profane, religious hate, political hate):
    - List evidence for classifying the sentence into this category.
    - List evidence against classifying the sentence into this category.
4. Summarize your findings and explain your final decision.
Your final output should be the category classification. Use only one of these exact category names: none, sexism, abusive, profane, religious hate, or political hate.

Provide your classification inside <classification> tags.